\documentclass{article} 
\usepackage{iclr2026_conference,times}

\usepackage{amsmath,amsfonts,bm}

\def\eqref#1{equation~\ref{#1}}

\def\1{\bm{1}}

\DeclareMathAlphabet{\mathsfit}{\encodingdefault}{\sfdefault}{m}{sl}
\SetMathAlphabet{\mathsfit}{bold}{\encodingdefault}{\sfdefault}{bx}{n}

\usepackage{hyperref}
\usepackage{url}
\usepackage[utf8]{inputenc} 
\usepackage[T1]{fontenc}    
\usepackage{booktabs}       
\usepackage{amsfonts}       
\usepackage{nicefrac}       
\usepackage{microtype}      
\usepackage{xcolor}         
\usepackage{microtype}
\usepackage{amsmath, amssymb}
\usepackage{booktabs}
\usepackage{multirow}
\usepackage{graphicx}
\usepackage{caption}
\usepackage{float}
\usepackage{enumitem}
\usepackage{hyperref}
\usepackage[most]{tcolorbox}

\newcommand{\pivot}{\textsc{PIVOT}}
\newcommand{\refine}{\textsc{REFINE}}
\newcommand{\model}[1]{\texttt{#1}}
\newcommand{\dataset}[1]{\textsc{#1}}
\newcommand{\Vone}{\textsc{V1}}
\newcommand{\Vtwo}{\textsc{V2}}
\newcommand{\Vthree}{\textsc{V3}}

\newcommand{\Fone}{\mathrm{F1}}
\newcommand{\Fonepivot}{\mathrm{F1}_{\pivot}}
\newcommand{\FoneRefine}{\mathrm{F1}_{\refine}}
\newcommand{\OverCarry}{\mathrm{OC}}

\title{Beyond Continuity: Challenges of Context Switching in Multi-Turn Dialogue with LLMs}

\author{%
  Aditya Sinha\thanks{corresponding author} \\
  Netflix Inc.\\
  Los Gatos, CA, USA \\
  \texttt{adityasinha@netflix.com} \\
  \And
  Harald Steck \\
  Netflix Inc.\\
  Los Gatos, CA, USA \\
  \texttt{hsteck@netflix.com} \\
  \And
  Vito Ostuni \\
  Netflix Inc.\\
  Los Gatos, CA, USA \\
  \texttt{vostuni@netflix.com} \\
  \And
  Matteo Rinaldi \\
  Netflix Inc.\\
  Los Gatos, CA, USA \\
  \texttt{matteorinaldi@netflix.com} \\
}

\newcommand{\meanstd}[2]{\ensuremath{#1{\scriptsize\,\pm\,}#2}}
\newcommand{\best}[1]{\textbf{\underline{#1}}}

\iclrfinalcopy 
\begin{document}

\maketitle

\begin{abstract}
  Users interacting with Large Language Models (LLMs) in a multi-turn conversation routinely \emph{refine} their requests or \emph{pivot} to new topics. LLMs, however, often miss these topic shifts and carry over irrelevant context from previous turns, leading to inaccurate responses. In this paper, we stress-test the multi-turn understanding of LLMs and study the following two sub-tasks: (1) detecting whether the user \emph{pivots} or \emph{refines} in the current turn, and (2) shortlisting relevant context from previous turns. To this end, we construct synthetic benchmarks based on real-world datasets from varied domains, as to simulate context shifts of different levels of difficulty. We then evaluate the \emph{zero-shot} performance of ten LLMs (open-weight, closed-source and reasoning), and demonstrate that only some reasoning and strongly instructed LLMs are accurate in detecting \emph{pivots}; open-weight LLMs struggle with the task and frequently carry stale context even with explicit \emph{cues}; and all models suffer from a position bias.
  Based on the results, we discuss key takeaways for improving long-term robustness in multi-turn capabilities for LLMs.
\end{abstract}

\section{Introduction}

In real-world dialogues with conversational assistants,  users routinely change topics mid-dialogue (i.e., without explicitly starting a new session). In contrast, LLM-based conversational assistants typically expect coherence and continuity in interactions, which leads to the carrying of stale information across turns. Deployment of \emph{off-the-shelf} models in chat based interfaces \cite{casheekar2024contemporary} without any task-specific fine-tuning is a realistic scenario for small enterprises, yet the ability of these models to \emph{reliably detect topic shifts} and \emph{reset context} at the \emph{current} user utterance in a multi-turn conversation is understudied in literature. To stress-test the multi-turn comprehension of the models, we frame this task as having two key aspects - (1) a classification problem predicting whether a user's intent is to \emph{pivot} to a new topic at the current turn (relative to the previous turns), or to \emph{refine} the current topic, and (2) a context selection task identifying prior turns that are relevant to the current turn. We evaluate these behaviours across several LLMs via prompt-based \emph{zero-shot} detection in controlled settings to probe the inherent generalization capability of out-of-the-box LLMs in multi-turn dialogue and expose their failure modes. 

Overall, our contributions are as follows:

\begin{itemize}
    \item We formalize the \emph{pivot} vs.\ \emph{refine} detection task together with appropriate \emph{context resetting} at \emph{pivots}, and construct synthetic conversations from real world datasets over diverse domains with precise control of pivot boundaries and their positions in a session.
    
    \item We introduce the \emph{over-carry rate} metric to quantify context carry-over errors, complementing standard $\Fone$ measures, offering a direct readout of \emph{reset} quality, as a useful evaluation metric for future work in assessing the quality of multi-turn understanding. 

    \item Since different real-world applications have varied latency/token constraints, we conduct an extensive \emph{zero-shot} prompt-based evaluation of ten LLMs spanning reasoning/non-reasoning and open/closed-weight families, providing (to our knowledge) one of the first direct comparisons of out-of-the-box models on \emph{pivot} detection and \emph{context resetting} capabilities.

    \item We analyze the impact of reasoning and cue phrases in the pivot detection and context resetting tasks. We also uncover how conversation structure with different pivot positions influences model performance, and offer insights into the failure modes for each model. This could be helpful for improving the performance of LLMs (and specifically open-weight models) for enhancing their multi-turn understanding. 

\end{itemize}

\section{Related Work}

Impressive progress has been made in conversational assistants' abilities to interact with human users in multi-turn conversations. Challenges like long-horizon decision-making or dealing with delayed rewards have been tackled by reinforcement learning (RL) in the context of multi-turn conversations with human users, e.g., \cite{zhou24, zhou25, jaques20,jang22},  as well as in the context of web-agents \cite{wei25,chae25,he24}. Maintaining  relevant context in memory is another challenge in multi-turn conversations, and has been addressed by query-rewriting in \cite{yang24,lai25} and memory augmentations in \cite{xu2021goldfishmemorylongtermopendomain,packer2024memgptllmsoperatingsystems, wang2023augmenting,zhong2024memorybank}. Humans often tend to start a conversation with under-specified queries, which was identified as an important unsolved problem in \cite{laban25}. Challenges and failure modes of LLMs for long-context management have been discussed in \cite{liu2023lost, lu2024insights}. A common underlying assumption in all these works is that the conversation follows a coherent topic, i.e., the user does not suddenly pivot to a different/unrelated topic during the conversation.

In real-world conversations, however, users routinely change topics suddenly \cite{soni2021empirical}. The task of topic shift detection in real-time dialogues was defined in \cite{xie21}:  \emph{Does the utterance of the user in the current turn of the conversation change the topic compared to the preceding turns?} Early methods to solve this problem used topic segmentation \cite{hearst1997text, eisenstein2008bayesian, galley2003discourse} and specialized classifiers \cite{konigari2021topic}, however they often struggled with incorporating subtle cues. Recent work uses hierarchical representations \cite{lin23algo} or task-specific methodology \cite{xie21, hwang24}  with smaller models, such as T5. We discuss additional related works in Section \ref{app:addrelatedwork} in the Appendix owing to space constraints. 

This serves as the primary motivation in our effort to construct a systematic study for stress-testing several recent LLMs in their \emph{inherent} ability to detect topic shifts in conversations via prompting alone, i.e. \emph{zero-shot} on base models without any task-specific fine-tuning, and expose their failure modes in detecting context switches.

\section{Problem Setting}

In this paper, we focus on a simple yet fundamental scenario: we assume that users either intend to \pivot{} from the topics discussed in the previous turn(s), or to \refine{} them further. Such a binary setup is very common in practice, as evidenced from prior research in real-world conversations \cite{soni2021empirical}. We then construct controlled ``hard switches" by concatenating head segments of different conversations, which also provides us with the ground-truth turn where the pivot(s) take place, as well as with the ground-truth shortlist of relevant prior turns.

Formally, let a conversation with T turns be defined by the sequence $\{(r_t, x_t)\}_{t=1}^{T}$ with role $r_t\in\{\textsc{user},\textsc{assistant}\}$ and utterance $x_t$. For each \textsc{user} turn $t$, we define:
\begin{align}
\text{Ground-truth label } \quad
y_t &\in \{\pivot, \refine\} \\
\text{Ground-truth shortlist of indices} \quad G_t &\subseteq \{1,\ldots,t-1\} \\
\text{Predicted label} \quad
\hat{y}_t &= f(x_{<t}, x_t) \in \{\pivot, \refine\} \\
\text{Predicted shortlist of indices} \quad
\hat{G}_t &= g(x_{<t}, x_t) \subseteq \{1,\ldots,t-1\}
\end{align}

where the shortlist of indices $G_t$ contains all the previous turns $t'<t$ that are relevant for the current turn $t$, based on the relevance of their utterances $x_{<t}$ to the current utterance $x_t$. It is easy to see that if the current turn is a pivot ($y_t=\pivot$), then $G_t=\varnothing$ . It is important to note that the predictions are functions ($f$ and $g$) of the current turn $t$ and the previous ones ($<t$), but not of future ones.

To evaluate the performance of a model, we report the 
classification metrics \(\Fonepivot\) and \(\FoneRefine\) computed from turn level predictions  $\hat{y}_t$ averaged
across all the conversations. While  \(\Fonepivot\) emphasizes a model’s sensitivity to true topic shifts (often sparse but consequential),  \(\FoneRefine\) tracks stability on the majority class. Reporting both the metrics mitigates any issues arising from class-imbalances. Further, \emph{context resetting} is evaluated from the predicted shortlist \(\hat{G}_t\), where at all the \emph{pivot} turns \(t\) in a conversation, the shortlist \(G_t=\varnothing\) (see above). Hence, we define the 
\emph{over-carry} rate of stale context at pivot-turns 
as 
\( \OverCarry = \text{mean }_{t\in\mathcal{I}_{\textsc{PIVOT}}}\mathbf{1}\{\,|\hat{G}_t|>0\,\}  \), where \(\mathcal{I}_{\textsc{PIVOT}}=\{t:\, y_t=\textsc{PIVOT}\}\) is the set of pivot-turns in the conversation (see Appendix Section \ref{app:metrics}). This metric directly probes whether a model can \emph{reset} its retrieval policy at a boundary, a frequently observed failure in long-context models \cite{liu2023lost}. Hence, the lower the value of $\OverCarry$ for a model, the better it is at \emph{resetting context}, and the smaller is the ``stickiness'' of stale context.

\section{Experiments}

In our experiments, we investigate the following three research questions: 
\textbf{RQ1:} How do the models perform on the tasks of PIVOT/REFINE detection and context resetting jointly? \textbf{RQ2}: How does the performance of the models for detection and context resetting change as the number of turns grow?, and finally \textbf{RQ3:} Do explicit context-switch signals and reasoning help in these tasks?  

\paragraph{Models and Scope:}

We evaluate three different classes of LLMs:

(1) open-weight models, (2) closed-source models, and (3) reasoning models -- please see Table \ref{tab:combined} for a detailed list.  We \emph{zero-shot} prompt the LLMs to generate a response at each turn $t$, comprised of 
a predicted label $\hat{y}_t \in\{\text{PIVOT},\text{REFINE}\}$ and a predicted shortlist $\hat{G}_t$ of relevant previous turns. (See Appendix Section \ref{app:modelsexperiments} for more details on the models, setup and prompts.). We focus on the task and evaluation for \emph{zero-shot} multi-turn understanding specifically, rather than downstream performance (See Section \ref{app:limitations}).

\paragraph{Dataset Construction:}
We leverage two datasets from different domains -- TopiOCQA \cite{adlakha2022topiocqa} with topic driven Q\&A, and MSC \cite{xu2021goldfishmemorylongtermopendomain} with \emph{persona} based chat conversations -- for constructing synthetic sessions by merging the head conversations from different conversations within the dataset. Specifically, we construct three settings -- 
\textbf{V1} with hard concatenation of conversations, \textbf{V2} with a \emph{cue-phrase} inserted at the pivot when concatenating conversations, making a pivot more natural and easier for the models to detect, and \textbf{V3} with a single pivot inserted at different positions (turn length) in the session. More details about the datasets, merging strategies and sample conversations can be found in Section \ref{app:datasets} in the Appendix.

\begin{figure*}[t]
  \centering
  \includegraphics[width=\textwidth]{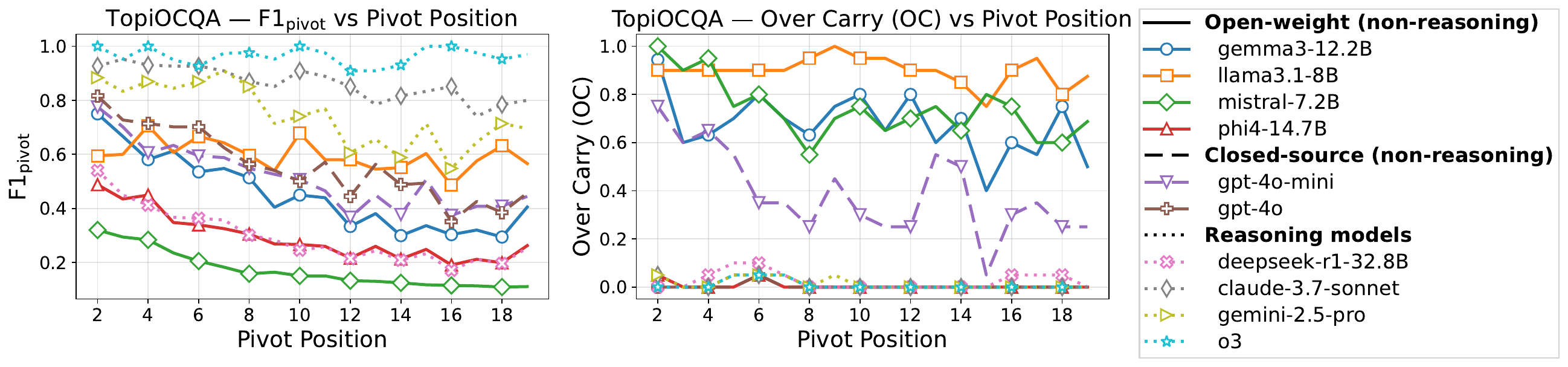}
  \caption{\dataset{TopiOCQA}: $\Fonepivot$ ( $\uparrow$ ) and $\OverCarry$ ($\downarrow$) vs. Pivot Position ($P$). Majority of the models show a degradation in $\Fonepivot$ as the Pivot Position increases, indicating challenging detection.}
  \label{fig:v4}
\end{figure*}

\paragraph{Results:}
Table~\ref{tab:combined} shows the results on the $\Vone$ and $\Vtwo$ settings for both datasets.
Furthermore, we report results on the $\Vthree$ setting of the TopiCOQA dataset in Figure \ref{fig:v4}, and discuss key findings below.

\begin{table}[h]
\centering
\caption{Results on \dataset{TopiOCQA} and \dataset{MSC} for \Vone{} (hard concat) and \Vtwo{} (cue) across families of models. Mean $\pm$ std over repeated runs for $\Fonepivot$, $\FoneRefine$, and pivot over-carry $\OverCarry$ (lower is better). We \textbf{highlight} the best result in each category and \underline{underline} the overall best result. }
\vspace{1em}
\label{tab:combined}
\small
\setlength{\tabcolsep}{4.5pt}
\resizebox{\linewidth}{!}{%
\begin{tabular}{l c c c c c c c}
\toprule
\multirow{2}{*}{\textbf{Model}} & \multirow{2}{*}{\textbf{Dataset}} &
\multicolumn{3}{c}{\textbf{TopiOCQA}} & \multicolumn{3}{c}{\textbf{MSC}} \\
\cmidrule(lr){3-5}\cmidrule(lr){6-8}
& & $\Fonepivot$~($\uparrow$) & $\FoneRefine$~($\uparrow$) & $\OverCarry$~($\downarrow$) &
      $\Fonepivot$~($\uparrow$) & $\FoneRefine$~($\uparrow$) & $\OverCarry$~($\downarrow$) \\
\midrule
\multicolumn{8}{l}{\textit{Open-weight (non-reasoning)}}\\
\addlinespace[3pt]
\multirow{2}{*}{\model{gemma3-12.2B}}      & \Vone & \meanstd{0.546}{0.001} & \meanstd{0.887}{0.002} & \meanstd{0.781}{0.003} & \meanstd{\bf{0.634}}{\bf{0.001}} & \meanstd{0.901}{0.002} & \meanstd{0.957}{0.001} \\
                                         & \Vtwo & \meanstd{0.591}{0.003} & \meanstd{0.908}{0.001} & \meanstd{0.837}{0.004} & \meanstd{\bf{0.652}}{\bf{0.002}} & \meanstd{0.906}{0.001} & \meanstd{0.929}{0.003} \\
\cmidrule(lr){1-8}
\multirow{2}{*}{\model{llama3.1-8B}} & \Vone & \meanstd{\bf{0.622}}{\bf{0.001}} & \meanstd{\bf{0.936}}{\bf{0.004}} & \meanstd{0.882}{0.001} & \meanstd{0.548}{0.001} & \meanstd{\bf{0.937}}{\bf{0.005}} & \meanstd{0.945}{0.002} \\
                                         & \Vtwo & \meanstd{\bf{0.699}}{\bf{0.007}} & \meanstd{\bf{0.946}}{\bf{0.002}} & \meanstd{0.877}{0.001} & \meanstd{0.774}{0.007} & \meanstd{\bf{0.960}}{\bf{0.002}} & \meanstd{0.927}{0.001} \\
\cmidrule(lr){1-8}
\multirow{2}{*}{\model{mistral-7.2B}}  & \Vone & \meanstd{0.214}{0.001} & \meanstd{0.180}{0.004} & \meanstd{0.756}{0.006} & \meanstd{0.295}{0.002} & \meanstd{0.380}{0.003} & \meanstd{0.991}{0.001} \\
                                         & \Vtwo & \meanstd{0.216}{0.003} & \meanstd{0.191}{0.001} & \meanstd{0.655}{0.009} & \meanstd{0.299}{0.004} & \meanstd{0.384}{0.001} & \meanstd{0.994}{0.009} \\
\cmidrule(lr){1-8}
\multirow{2}{*}{\model{phi4-14.7B}}     & \Vone & \meanstd{0.367}{0.003} & \meanstd{0.731}{0.001} & \meanstd{\bf{0.002}}{\bf{0.004}} & \meanstd{0.486}{0.002} & \meanstd{0.791}{0.003} & \meanstd{\bf{0.248}}{\bf{0.001}} \\
                                         & \Vtwo & \meanstd{0.385}{0.003} & \meanstd{0.756}{0.006} & \meanstd{\bf{0.003}}{\bf{0.004}} & \meanstd{0.503}{0.002} & \meanstd{0.808}{0.002} & \meanstd{\bf{0.224}}{\bf{0.005}} \\
\addlinespace[2pt]
\midrule
\multicolumn{8}{l}{\textit{Closed-source (non-reasoning)}}\\
\addlinespace[3pt]
\multirow{2}{*}{\model{gpt-4o-mini}}     & \Vone & \meanstd{0.612}{0.002} & \meanstd{0.915}{0.001} & \meanstd{0.446}{0.010} & \meanstd{\bf{0.812}}{\bf{0.003}} & \meanstd{\bf{0.964}}{\bf{0.002}} & \meanstd{1.000}{0.000} \\
                                         & \Vtwo & \meanstd{0.645}{0.005} & \meanstd{0.927}{0.002} & \meanstd{0.426}{0.006} & \meanstd{\bf{0.839}}{\bf{0.005}} & \meanstd{\bf{0.969}}{\bf{0.002}} & \meanstd{0.970}{0.006} \\
\cmidrule(lr){1-8}
\multirow{2}{*}{\model{gpt-4o}}          & \Vone & \meanstd{\bf{0.695}}{\bf{0.007}} & \meanstd{\bf{0.943}}{\bf{0.002}} & \meanstd{\bf{0.002}}{\bf{0.000}} & \meanstd{0.618}{0.007} & \meanstd{0.889}{0.002} & \meanstd{\bf{0.036}}{\bf{0.001}} \\
                                         & \Vtwo & \meanstd{\bf{0.708}}{\bf{0.009}} & \meanstd{\bf{0.947}}{\bf{0.002}} & \meanstd{\bf{0.000}}{\bf{0.000}} & {\meanstd{0.635}{0.009}} & {\meanstd{0.896}{0.002}} & \best{\meanstd{\bf{0.015}}{\bf{0.001}}} \\
                                
\addlinespace[2pt]
\midrule
\multicolumn{8}{l}{\textit{Reasoning models}}\\
\addlinespace[3pt]
\multirow{2}{*}{\model{deepseek-r1-32.8B}} & \Vone & \meanstd{0.373}{0.003} & \meanstd{0.740}{0.001} & \meanstd{0.029}{0.002} & \meanstd{0.497}{0.004} & \meanstd{0.804}{0.001} & \meanstd{0.203}{0.002} \\
                                         & \Vtwo & \meanstd{0.408}{0.002} & \meanstd{0.783}{0.002} & \meanstd{0.011}{0.001} & \meanstd{0.535}{0.005} & \meanstd{0.833}{0.004} & \meanstd{0.094}{0.001} \\
\cmidrule(lr){1-8}
\multirow{2}{*}{\model{claude-3.7-sonnet}}     & \Vone & \meanstd{0.941}{0.003} & \meanstd{0.992}{0.002} & \meanstd{0.006}{0.003} & \meanstd{0.703}{0.002} & \meanstd{0.927}{0.004} & \meanstd{0.109}{0.003} \\
                                         & \Vtwo & \meanstd{0.945}{0.002} & \meanstd{0.992}{0.001} & \meanstd{0.001}{0.003} & \meanstd{0.717}{0.002} & \meanstd{0.931}{0.004} & \meanstd{0.051}{0.002} \\
\cmidrule(lr){1-8}
\multirow{2}{*}{\model{gemini-2.5-pro}}     & \Vone & \meanstd{0.871}{0.003} & \meanstd{0.981}{0.002} & \meanstd{0.004}{0.002} & \meanstd{0.713}{0.004} & \meanstd{0.931}{0.001} & \meanstd{\bf{0.048}}{\bf{0.004}} \\
                                         & \Vtwo & \meanstd{0.879}{0.002} & \meanstd{0.982}{0.003} & \meanstd{0.001}{0.001} & \meanstd{0.725}{0.004} & \meanstd{0.934}{0.003} & \meanstd{\bf{0.024}}{\bf{0.002}} \\

\cmidrule(lr){1-8}
\multirow{2}{*}{\model{o3}}              & \Vone & \meanstd{\bf{0.973}}{\bf{0.003}} & \meanstd{\bf{0.997}}{\bf{0.002}} & \meanstd{\bf{0.003}}{\bf{0.001}} & \meanstd{\bf{0.763}}{\bf{0.003}} & \meanstd{\bf{0.965}}{0.004} & \meanstd{0.309}{0.002} \\
                                         & \Vtwo & \best{\meanstd{\bf{0.977}}{\bf{0.001}}} & \best{\meanstd{\bf{0.997}}{\bf{0.001}}} & \best{\meanstd{\bf{0.000}}{\bf{0.002}}} & \best{\meanstd{\bf{0.861}}{\bf{0.003}}} & \best{\meanstd{\bf{0.977}}{0.002}} & \meanstd{0.176}{0.001} \\

\bottomrule
\end{tabular}
}
\end{table}

\paragraph{RQ1: Open-weight models are \emph{sticky}, Closed-source models perform better.} \model{gpt-4o} dominates the non-reasoning baselines, combining high $\Fonepivot$ and high $\OverCarry$, followed by \model{gpt-4o-mini} which is competitive on detection but \emph{over-carries} dramatically, revealing a strong continuity bias. Amongst open weight models, two clear failure modes emerge evidently, where the first ones ``always-carry'' the context, such as \model{gemma3-12.2B} and \model{llama-3.1-8B} which show reasonable $\Fonepivot$ but very high $\OverCarry$ as seen on \dataset{TopiOCQA} and more so on \dataset{MSC}. They exhibit high $\OverCarry$ even when they predict $\pivot$, i.e. they acknowledge that the conversation is \emph{pivoting} but still carry unrelated context. This indicates that their ``classification-head'' and ``context selection'' are potentially not well aligned, possibly requiring further instruction tuning. The second failure mode which is evident, is the \emph{low/empty} carrying of the context, for instance \model{phi4-14.7B}, which shows very low $\OverCarry$, but also a lower $\Fonepivot$ indicating a conservative context resetting, but weak detection pattern. \model{mistral-7.2B} underperforms on both $\Fonepivot$ and $\OverCarry$ in our settings.

\paragraph{RQ2: Position bias of context.} Figure \ref{fig:v4} shows a trend of degraded $\Fonepivot$ performance across models as the pivot position increases, indicating that \emph{later} pivots in a session become much harder to detect, while $\OverCarry$ fluctuates. This reveals a clear \textbf{position bias} and complements the unreliability of LLMs as discussed in \cite{laban25}. All non-reasoning models show a rapid downward trend, indicating that instruction following is not sufficient for this task, while reasoning models show a more gradual decline indicating that while an ``active'' state reset via reasoning helps detection, reasoning models still struggle with this task as the number of turns grow. 

\paragraph{RQ3a: Cue phrases help label prediction.} We observe that $\Vtwo$ with \emph{cue-bridged} pivots improves $\Fonepivot$ and in some instances $\OverCarry$ also across all models, compared to $\Vone$.  This indicates that the \emph{cue-phrases} likely provide a ``lexicalized boundary token'' which gives a clear pivoting signal to the models.
We observe that models with strong instruction following ability (\model{gpt-4o}, \model{gpt-4o-mini}) benefit more from these cues than models where instruction tuning is relatively weaker. 

\textbf{RQ3b: Reasoning acts as ``gating'' context.} \model{o-3} achieves high $\Fonepivot$ on \dataset{TopiOCQA} with zero $\OverCarry$ in V2, and competitive performance on \dataset{MSC} with reduced but non-zero $\OverCarry$, followed by \model{claude-3.7-sonnet}, \model{gemini-2.5-pro} and \model{deepseek-r1-32.8B} in performance order. The low $\OverCarry$ across these models indicates that reasoning leads to a calibrated ``gating'' of context, i.e. the model drops unrelated context at topic switches, while non-reasoning models fail at the task.

\vspace{-0.3em}
\section{Conclusions and Future Work}
In this work, we stress-test the multi-turn capabilities of LLMs and expose the challenges in \emph{pivot-detection} and \emph{context resetting} for practical scenarios. 
Our main findings are that a subset of the LLMs with strong instruction-following perform reasonably on detection of \emph{pivots} but struggle when the switches occur at later turns; majority of the state-of-the-art open-weight LLMs often get confused as to where a \emph{pivot} takes place and carry over stale context; and explicit reasoning, understandably, is capable of mitigating some of the challenges around \emph{pivot-detection}. There are several solutions worth exploring for improving the multi-turn capabilities of these models, such as \emph{targeted} fine-tuning, \emph{two-stage} pipelining, \emph{position-aware} context decaying and \emph{inference-time guards}. We discuss these methods with model specific failures for the benefit of practitioners, and limitations in Sections \ref{app:adddiscusion} and \ref{app:limitations} of the Appendix respectively, and defer these to future work.  

\subsubsection*{Acknowledgments}

We thank our colleagues and collaborators Becks Wood,  Christopher Huang, Claire Campbell, Ehsan Gholami, Grace Huang, Guru Tahasildar, Hakan Baba,  Ivan Provalov, Jeremey Fleischer, Jining Zhong,  Kelley Robinson, Kyle Fox, Moumita Bhattacharya, Rhodes Kelley, Shahrzad Naseri, Shreyas Suhas Chavan, Spencer L'Heureux, Sudarshan Lamkhede, Thea Wang,  Veli Balin,  Vicky Liu, Yesu Feng, Yibo Dai and  Zhe Zhang for helpful discussions and feedback throughout this project. We are grateful to the workshop organizers and reviewers for their time and constructive comments. This work was supported by internal funding and computing resources provided by Netflix Inc.

\bibliography{iclr2026_conference}
\bibliographystyle{iclr2026_conference}

\clearpage
\appendix
\section*{Appendix}

This appendix is segmented into the following key parts.

\begin{enumerate}

    \item \textbf{Section \ref{app:addrelatedwork}} discusses additional related works in the domain of topic-switching, modeling and multi-turn interactions using LLMs not included in the main paper due to lack of space. 

    \item \textbf{Section \ref{app:datasets}} and \textbf{Section \ref{app:experiments}} provide details about the construction and use of datasets, experimental details about the evaluation, metrics, and setup. 

    \item \textbf{Section \ref{app:adddiscusion}} provides an additional discussion on the results, identifying the role of the dataset domain, identifying model-specific failures and offering recommendations for improving the capabilities of models for multi-turn interactions. 

    \item \textbf{Section \ref{app:limitations}} discusses the current scope of this study and limitations that we recognize in the analysis.

    \item \textbf{Section \ref{app:statements}} mentions the Ethics Statement, Reproducibility Statement, and Statement on LLM Usage for our research. 
    
\end{enumerate}

\section{Additional Related Work}\label{app:addrelatedwork}

\paragraph{Topic Shift Detection and Topic Segmentation.}   \cite{hearst1997text} proposes a lexical cohesion method for detecting topic boundaries by establishing foundational segmentation signals, while \cite{galley2003discourse} extends segmentation to conversations using a combination of lexical cohesion and prosodic cues. \cite{eisenstein2008bayesian} introduces a bayesian model for topic boundary discovery to improve robustness in noisy conversational data, and \cite{xie21} frames topic-shift aware dialog modeling as a separate task. \cite{lin23algo} uses prompt-based learning to capture topic signals and \cite{hwang24} generates dialogues with natural topic transitions via knowledge-graph paths. 

\paragraph{Topic Aware Dialogue Modeling.} \cite{xu21} segments context into topic-aware units and matches them to candidate responses to improve response selection with topic shifts, \cite{feng2022topic} conditions generation on topical signals to better ground responses in the right knowledge, and \cite{ma2024multi} proposes unsupervised topic-shift detection to segment dialogues and improve downstream tasks.

\paragraph{Long-Context and Multi-Turn Reliability.} \cite{liu2023lost} demonstrates that models under-use information in the middle of long contexts, whereas \cite{lu2024insights} performs probing analysis to reveal that models often encode needed information but fail to use it in the outputs, motivating explicit context management. \cite{laban25} benchmarks various LLMs across multi-turn tasks and find that reliability drops and recovery becomes poor when the models make early mistakes. 

\paragraph{Task Oriented Assistants and Context Carryover.} \cite{chen2019improving} jointly models candidate slots to decide the context to be carried over across turns for speech systems, while \cite{naik2018contextual} proposes a neural decision framework for deciding carry slot values across different domains. \cite{wanigasekara2022multimodal} extends carryover to combined voice and vision use cases with system modifications. 

\paragraph{Conversational Memory.} \cite{xu2021goldfishmemorylongtermopendomain} releases multi-session dialogue data and show that retrieval-augmented and summarization based memory helps long horizon conversations. \cite{packer2024memgptllmsoperatingsystems} implements hierarchical ``virtual context'' with interrupts and memory tiers to manage unbounded conversational history, and \cite{wang2023augmenting} adds a decoupled memory reader with a frozen backbone for recalling long-term context efficiently. \cite{zhong2024memorybank} proposes a persistent memory module with decay and refresh to maintain user aligned facts across sessions.

\section{Datasets}\label{app:datasets}

\subsection{Raw datasets}\label{app:raw-datasets}

\paragraph{TopiOCQA.}
We use \dataset{TopiOCQA} (\emph{Topical Open-domain Conversational QA}) from \cite{adlakha2022topiocqa}, a multi-turn QA corpus where utterances are grounded in Wikipedia passages. The domain is factual, entity-centric QA with short system responses and information-seeking user turns. We pool the train/dev/test splits and retain only English sessions.\footnote{\href{https://github.com/McGill-NLP/topiocqa}{https://github.com/McGill-NLP/topiocqa}} In our benchmarks, \dataset{TopiOCQA} serves as a \emph{knowledge-seeking} domain where pivots are typically signalled by lexical shifts (new entities, events, or facets), and refinement is characterized by follow-ups on the same entry.

\paragraph{MSC }
We also use \dataset{MSC} (Multi-Session Chat) from \citep{xu2021goldfishmemorylongtermopendomain}, which comprises long-horizon conversations between human personas over multiple sessions. Dialog turns are free-form, less entity-anchored, and frequently rely on pragmatic continuity (speaker goals, preferences, commitments). Relative to \dataset{TopiOCQA}, \dataset{MSC} is a \emph{social/dialogue} domain where pivots are subtler (persona shifts or goal changes) and lexical cues are weaker, making context discipline harder to learn.\footnote{\href{https://parl.ai/projects/msc/}{https://parl.ai/projects/msc/}}

\paragraph{Licensing and filtering.}
We respect dataset licenses. No additional annotation beyond our pivot/refine construction is performed on the raw corpora (see Appendix~\ref{app:limitations} for scope limits).

\subsection{Overview of synthetic benchmarks}\label{app:synthetic-overview}

We construct three stress-test variants that introduce \emph{controlled} pivots by concatenating head segments from different source conversations. Each variant preserves turn-order and speaker roles and provides a \emph{ground-truth shortlisted context} per user turn (indices of prior messages considered relevant to answer the current turn).

\begin{itemize}[leftmargin=1.2em]
\item \textbf{V1 — Hard concatenation (no cue).} Two conversations are concatenated: the first $k$ user turns from conversation $A$, then the first $p$ user turns from conversation $B$. The very first turn of $B$ is labeled \textsc{PIVOT}; all others are \textsc{REFINE}. No bridging text is inserted.
\item \textbf{V2 — Cue-bridged pivot.} Same as V1 but we prepend a short cue phrase (e.g., ``switching gears,'' ``on a different note,'') at the pivot user turn to provide explicit lexical indication of the shift.
\item \textbf{V3 — Pivot position study.} We vary the pivot user-turn position $P\in[2,20]$ within a session to quantify position bias. Each session has exactly one pivot at user turn $P$.
\end{itemize}

To avoid leakage and topic recurrence, a source conversation ID appears at most once per synthetic session.

\subsubsection{V1: Construction details and examples}\label{app:v1-details}

\paragraph{Generation recipe.}
Given a pool of source conversations $\mathcal{C}$, we sample two distinct IDs $A,B\in\mathcal{C}$. We take the first $k$ user turns from $A$ and the first $p$ user turns from $B$, respecting original (user/system) alternation. We label the very first user turn from session $B$ as a \textsc{PIVOT} and all other user turns as \textsc{REFINE}. The \emph{ground-truth shortlisted context} for a user turn $t$ is defined as the set of prior turn indices from the \emph{same} segment (e.g., all earlier messages from $A$ if $t\in A$). For a pivot turn (first turn from $B$), the shortlist is empty by construction.

\paragraph{Sampling hyperparameters.}
We use randomly sampled $k\in\{2,3,4\}$ and $p\in\{1,\ldots,6\}$ (task-dependent), subject to a session budget of up to 20 total utterances. We enforce segment uniqueness (each $(A,B)$ pair used at most once per split). Exact values are reported in Table~\ref{tab:synth-stats}.

We show two V1 examples (user turns are marked with \textsc{REFINE}/\textsc{PIVOT} and the ground-truth shortlist indices for that turn):
\paragraph{Examples (TopiOCQA).}

\begin{small}
\begin{verbatim}
=== Synthetic Session v1grid_rr_000000 [v1_hard_pivot_no_cue] ===
Segments: [{'conv_id': '289', 'head_user_turns': 2}, 
           {'conv_id': '815', 'head_user_turns': 5}]
Num pivots: 1 | Pivot user-turn positions: [3]
Variant params: {'k_users_A': 2, 'p_users_B': 5, 'user_turn_budget': 20}
------------------------------------------------------------------------------------------
U01[REFINE] (seg=1, src:289@0): who are the original avengers in the movies
S01 (seg=1, src:289@1): tony stark, steve rogers, bruce banner, and thor.
U02[REFINE] (seg=1, src:289@2): who directed it
   -> ctx[0|user] who are the original avengers...
   -> ctx[1|system] tony stark, steve rogers...
S02 (seg=1, src:289@3): joss whedon
U03[PIVOT] (seg=2, src:815@0): how long do painter turtle eggs take to hatch
...
\end{verbatim}
\end{small}

\paragraph{Examples (MSC).}
\begin{small}
\begin{verbatim}
=== Synthetic Session v1grid_rr_000001 [v1_hard_pivot_no_cue] ===
Segments: [{'conv_id': 'train:..._7529::session_3', 'head_user_turns': 4},
           {'conv_id': 'train:..._3782::session_3', 'head_user_turns': 4}]
Num pivots: 1 | Pivot user-turn positions: [5]
Variant params: {'k_users_A': 4, 'p_users_B': 4, 'user_turn_budget': 12}
------------------------------------------------------------------------------------------
U01[REFINE] ... The weather was so nice today so I decided to bike to work.
S01 ... Good for you! Was it blue skies or did you have some nice clouds...
...
U05[PIVOT] ... My parents just asked me if we had any baked goods...
...
\end{verbatim}
\end{small}

\paragraph{Dataset statistics.}
Table~\ref{tab:synth-stats} summarizes V1 statistics per source corpus (sessions, user turns, mean/min/max turns).

\begin{table}[H]
\centering
\caption{Synthetic dataset statistics}
\label{tab:synth-stats}
\small
\begin{tabular}{lrrrr}
\toprule
Variant \& Source & \#Sessions & \#User turns & Mean turns & Min/Max turns\\
\midrule
V1–TopiOCQA & \texttt{625} & \texttt{5700} & \texttt{10.12} & \texttt{<4/18>}\\
V1–MSC      & \texttt{330} & \texttt{2340} & \texttt{8.09} & \texttt{<4/12>}\\
\bottomrule
\end{tabular}
\end{table}

\subsubsection{V2: Cue-bridged pivot and examples}\label{app:v2-details}

\paragraph{Generation recipe.}
V2 mirrors V1 but we insert a lexical cue at the pivot user turn. Let $\mathcal{C}_{\text{cue}}$ be a curated list of brief phrases from the exact list [``switching gears,'' ``on a different note,'' ``new topic,'' ``separately,'', ``switching the topic,'' ``changing subject,'' ``another thing,''] sampled uniformly and prepended to the pivot utterance. All other labels and shortlisted context rules remain identical to V1, and hence shares the same dataset statistics as shown in Table \ref{tab:synth-stats}. Cues are lower-cased and punctuated minimally to avoid biasing the language model toward certain domains.

\paragraph{Example (TopiOCQA).}
\begin{small}
\begin{verbatim}
=== Synthetic Session v2grid_rr_000000 [v2_hard_pivot_with_cue] ===
...
U03[PIVOT*CUE] ... switching gears, how long do painter turtle eggs take to hatch
...
\end{verbatim}
\end{small}

\paragraph{Example (MSC).}
\begin{small}
\begin{verbatim}
=== Synthetic Session v2grid_rr_000001 [v2_hard_pivot_with_cue] ===
...
U05[PIVOT*CUE] ... on a different note, My parents just asked me if we had 
any baked goods...
...
\end{verbatim}
\end{small}

\subsection{V3: Pivot position study}\label{app:v3-details}

\paragraph{Generation recipe.}
We vary the pivot position $P$ in a single-pivot synthetic session. For each synthetic session we first sample a base conversation $A$ and take sufficient head turns to admit a pivot at user turn $P$. We then splice in the first $p$ turns from a distinct conversation $B$ starting at that position. The turn at $P$ is labeled \textsc{PIVOT}; all earlier turns ($\leq P$) are \textsc{REFINE} and all later turns are \textsc{REFINE}. Ground-truth shortlists follow the segment-local rule; the pivot shortlist is empty.

\paragraph{Examples and plots.}
See Fig.~\ref{fig:v4} in the main paper for \dataset{TopiOCQA} V3 trends ($\Fonepivot$ and pivot over-carry vs.\ $P$), and Appendix~\ref{app:limitations} for discussion of position-bias implications.

\subsection{Pre-processing and quality control}\label{app:qc}

We normalize whitespace and strip system templates not part of the original corpora. We enforce that conversation IDs are not reused within a synthetic session and only appear once to prevent topical leakage. All synthetic creation retains the original role order.

\subsection{Why synthetic composition?}
Naturally occurring pivots in public corpora are sparse and ambiguous. Moreover, annotating turn-level pivot/refine at scale requires non-trivial annotation efforts, and also results in inter-annotator differences. In contrast, our controlled concatenations provide the ground-truth labels PIVOT/REFINE by construction. This allows us to evaluate a model’s \emph{reset policy} and \emph{context discipline} under varied conditions (explicit cues and pivot positions) while keeping the underlying language realistic (sourced from \dataset{TopiOCQA} and \dataset{MSC}). This complements prior work on session memory and topical drift with a diagnostic lens focused on \emph{boundary detection} and \emph{over-carry}. \cite{naik2018contextual, chen2019improving}

\section{Experiments}\label{app:experiments}

\subsection{Evaluation metrics}\label{app:metrics}

We evaluate two complementary capabilities that underlie robust multi-turn dialogue: (i) \emph{boundary detection}—recognizing whether a user turn \emph{pivots} to a new topic or \emph{refines} the current one; and (ii) \emph{context discipline}—shortlisting only the prior turns relevant to answer the current request without spuriously carrying stale context forward. The first is measured with standard classification metrics; the second is captured primarily by an \emph{over-carry} rate at pivot turns. This design follows prior work on carry-over detection in long-context dialogue systems \cite{chen2019improving}.

\paragraph{Setup and notation.}
Let each user turn \(t\) have a ground-truth label \(y_t \in \{\textsc{PIVOT}, \textsc{REFINE}\}\) and a model prediction \(\hat{y}_t\) from the same set. For shortlist evaluation, let \(G_t \subset \{1,\dots,t-1\}\) be the \emph{ground-truth} set of prior turn indices deemed relevant for turn \(t\), and let \(\hat{G}_t\) be the model’s predicted set. In our synthetic benchmarks, \(G_t=\varnothing\) for pivot turns by construction; for refine turns \(G_t\) consists of within-segment history up to \(t{-}1\) (See Appendix~\ref{app:datasets} for details of construction).

\subsubsection*{Classification metrics: pivot vs.\ refine}

We report per-class $\Fone$ for both \(\textsc{PIVOT}\) and \(\textsc{REFINE}\). For a class \(c\in\{\textsc{PIVOT},\textsc{REFINE}\}\),
\[
\text{Prec}_c
= \frac{\mathrm{TP}_c}{\mathrm{TP}_c + \mathrm{FP}_c},\quad
\text{Rec}_c
= \frac{\mathrm{TP}_c}{\mathrm{TP}_c + \mathrm{FN}_c},\quad
\Fone_c
= \frac{2\,\text{Prec}_c\,\text{Rec}_c}{\text{Prec}_c + \text{Rec}_c}.
\]

We denote $\Fone_c$ for $\pivot$ as  \(\Fone_{\textsc{PIVOT}}\) and $\Fone_c$ for $\refine$ \(\Fone_{\textsc{REFINE}}\).

\emph{Motivation.} \(\Fonepivot\) emphasizes a model’s sensitivity to true topic shifts (often sparse but consequential), while \(\FoneRefine\) tracks stability on the majority class in many real conversations. Reporting both mitigates class-imbalance artifacts and exposes asymmetries such as ``always-refine'' or ``always-pivot'' behaviors.

\subsubsection*{Context discipline: carryover metrics}

\paragraph{Pivot over-carry.} 
At a true topic switch, the expected behavior for a model is to drop the old context and start fresh. Hence, if the model is still bringing along context from any earlier turns, it can be said to be \emph{over-carrying} i.e., allowing stale context to carry-over into the new topic, when it should have been rightfully \emph{reset}. At a pivot turn \(t\), the ground-truth shortlist is empty by design (\(G_t=\varnothing\)). Any non-empty predicted shortlist is an \emph{over-carrying} error. We define the \emph{over-carry} rate in an conversation as 
\[
\OverCarry \;=\; \frac{1}{|\mathcal{I}_{\textsc{PIVOT}}|}\sum_{t\in\mathcal{I}_{\textsc{PIVOT}}}\mathbf{1}\{\,|\hat{G}_t|>0\,\},
\]
where \(\mathcal{I}_{\textsc{PIVOT}}=\{t:\, y_t=\textsc{PIVOT}\}\).
Lower is better. This metric directly probes whether the model can \emph{reset} its retrieval policy at a boundary, a failure mode frequently observed in long-context models \cite{lu2024insights}.

\paragraph{Why we emphasize pivot over-carry.}
Over-carry at pivot is \emph{well-identified} in our synthetic design (ground-truth shortlist is provably empty), making it a robust diagnostic for reset behavior without needing any human intervention. In contrast, \emph{under-carry}, i.e. failing to retrieve enough prior context on refine turns is harder to measure faithfully without curated relevance labels: not all within-segment messages are strictly necessary.

For this reason, we treat \(\OverCarry\) at pivot as the primary signal of context discipline.

\paragraph{Reporting and aggregation.}
Unless noted, we average across all user turns within a split and report per-model/per-variant means \(\pm\) standard deviations over 2 repeated runs (Table~\ref{tab:combined}). For V3 we additionally condition by pivot position \(P\) to analyze position bias in both \(\Fonepivot\) and \(\OverCarry\) (Fig.~\ref{fig:v4}).

\subsection{Models and Setup}\label{app:modelsexperiments}

\paragraph{Models.}
We evaluate ten models  spanning open/closed weights and reasoning vs.\ standard instruction tuning:
\begin{itemize}[leftmargin=1.2em]
  \item \textbf{Closed, non-reasoning:} \model{GPT\textendash 4o} \cite{openai2024_gpt4o_systemcard}, \model{GPT\textendash 4o-mini} \citep{openai2024_gpt4omini_blog}, 
  \item \textbf{Closed, reasoning:} \model{Gemini-2.5-pro} \cite{deepmind2025_gemini25pro}, \model{Claude-3.7-sonnet} \cite{anthropic2025_claude37sonnet}, \model{OpenAI o3} \cite{openai2025_o3_systemcard}
  \item \textbf{Open, non-reasoning:} \model{Llama\,3.1\textendash8B} \cite{meta2024_llama31_8b}, \model{Gemma\,3\textendash12.2B} \cite{google2025_gemma3_12b}, \model{Phi-4\textendash14.7B} \cite{microsoft2024_phi4_14b}, \model{Mistral\textendash7.2B} \cite{mistral2024_mistral7b_v03}
  \item \textbf{Open, reasoning:} \model{DeepSeek-R1\textendash32.8B} \cite{deepseek2025_r1_32b}
\end{itemize}
We keep official model defaults unless stated and treat vendor hosted models as black boxes (no weight access). 
\paragraph{Hardware and inference environment.}
Closed models are accessed via the provider’s hosted inference endpoints (region and provisioning managed by the vendor). Open-weight models are served locally with \texttt{ollama} on a Linux host with Nvidia A100x4 GPUs with 40 GB memory. All experiments are executed under fixed seeds for reproducibility. Code and instructions to download and process the datasets for the paper can be found at the URL: \href{https://github.com/adityaasinha28/BeyondContinuityLLMs}{\texttt{https://github.com/adityaasinha28/BeyondContinuityLLMs}}. 

\paragraph{Licensing and usage.}
We respect each model’s license: closed models are accessed under their ToS; open-weight models (Llama, Gemma, Mistral, Phi-4, DeepSeek R1) are used for \emph{research benchmarking only}. We do not fine-tune or adapt any model in this study.

\paragraph{Prompting and JSON conformance.}
All models receive a single \emph{joint} prompt per user turn (history $\to$ current message) with a compact system instruction requiring \emph{JSON-only} output in this format:
\[
\{\texttt{answer}:\text{str},\;\texttt{predicted\_label}:\{\textsc{PIVOT},\textsc{REFINE}\},\;\texttt{relevant\_context}:\text{int[]}\}.
\]
We use a strict parser with guardrails that rejects schema with incorrect formats and additional keys. For open-weight models we apply a minimal JSON prefix/suffix fence and stop tokens when available. We skip the first user turn in every session for both inference and evaluation (as it has no usable history). We use API calls with retries for instances with schema violation, although we did not observe any significant violation rates across the models. 

\paragraph{Decoding and runtime.}
For vendor-hosted models (GPT-4o/\-mini, o3, gemini-2.5-pro, claude-3.7-sonnet) we use \(\texttt{temperature}{=}0.0\) with \(\texttt{reasoning}\) parameter set, wherever applicable. 
For open-weight models via \texttt{ollama} we use:
\[
\texttt{temperature}=0.0,\quad
\texttt{top\_p}=0.9,\quad 
\texttt{top\_k}=50,\quad
\]

\subsection{Prompts}\label{app:prompts}

\begin{tcolorbox}[
  title={\textbf{System prompt --- TopiOCQA}},
  breakable,
  enhanced,
  colback=gray!2,
  colframe=black!12,
  arc=2pt,
  outer arc=2pt,
  listing engine=listings,
  listing only,
  listing options={style=promptstyle}
]
You are a helpful assistant interacting with the user.

Given the conversation history and the CURRENT user message, your task is to:

1) Respond to the user with an ANSWER (concise, <= 40 words).

2) Classify the CURRENT message as PIVOT (switching to a new topic) or REFINE (continuing on the same thread).

3) Select which PRIOR message ids (ints) are relevant context to answer the CURRENT message.

STRICT OUTPUT FORMAT RULES:

- Return ONE JSON object only (no prose, no markdown, no code fences).

- Use EXACTLY these keys: answer, predicted\_label, relevant\_context.

- Valid predicted\_label values must be one of these: "PIVOT" or "REFINE".

- relevant\_context must be an array of integers (ids from the history).

- Do NOT include any extra keys like "type", "properties", "required", "schema", or examples.

- Do NOT include the current message id in relevant\_context.

- Do NOT make up ids that are not present in the history.

Your output must be valid JSON matching:

\{ "answer": "<string>", "predicted\_label": "PIVOT"|"REFINE", "relevant\_context": [<int>, ...] \}
\end{tcolorbox}

\vspace{0.75em}

\begin{tcolorbox}[
  title={\textbf{System prompt — MSC (Persona Conversations)}},
  listing only,
  breakable,
  enhanced,
  colback=gray!2,
  colframe=black!12,
  arc=2pt,
  outer arc=2pt,
  boxsep=1ex,
  left=1ex, right=1ex, top=1ex, bottom=1ex
]
You are a helpful assistant interacting with the user.

Given the conversation history and the CURRENT user message, your task is to:

1) Respond to the user with an ANSWER (concise, <= 40 words).

2) Classify whether the CURRENT user message persona is PIVOT (switching to a new topic and persona) or REFINE (continuing on the same thread and persona).

3) Select which PRIOR message ids (ints) are relevant context and consistent persona with the CURRENT message.

STRICT OUTPUT FORMAT RULES:

- Return ONE JSON object only (no prose, no markdown, no code fences).

- Use EXACTLY these keys: answer, predicted\_label, relevant\_context.

- Valid predicted\_label values must be one of these: "PIVOT" or "REFINE".

- relevant\_context must be an array of integers (ids from the history).

- Do NOT include any extra keys like "type", "properties", "required", "schema", or examples.

- Do NOT include the current message id in relevant\_context.

- Do NOT make up ids that are not present in the history.

Your output must be valid JSON matching:

\{ "answer": "<string>", "predicted\_label": "PIVOT"|"REFINE", "relevant\_context": [<int>, ...] \}
\end{tcolorbox}

\section{Additional Discussion on Results}\label{app:adddiscusion}

\subsection{Dataset and Domain}
The domain of the underlying dialogue shapes both the target behavior and the dominant error modes. On \dataset{TopiOCQA} (knowledge-seeking, task-driven QA), models are implicitly rewarded for \emph{resetting} context when the topic changes and for grounding answers in the most recent, local evidence. This inductive bias aligns well with our pivot/shortlist objectives: most systems achieve higher $\Fonepivot$ and lower over-carry (OC) on \dataset{TopiOCQA} than on \dataset{MSC}. In contrast, \dataset{MSC} is persona-centric chit-chat, where human turns are stylistically and semantically similar across long spans. Models trained with conversational objectives and RLHF priors that emphasize politeness, empathy, and “follow the thread” coherence exhibit a strong \emph{continuity prior}: even when they correctly flag a PIVOT, they often continue to retrieve or reference stale persona turns, showing higher $\OverCarry$.

These domain-conditioned behaviors are visible across families. Closed-source non-reasoning models (e.g., \model{gpt-4o}, \model{gpt-4o-mini}) tend to be context-disciplined on \dataset{TopiOCQA}—near-zero OC when a pivot is detected—yet show stickiness on \dataset{MSC}, where the conversational reward model has likely learned that maintaining continuity is helpful. Open-weight mid-size models (\model{gemma3-12B}, \model{llama3.1-8B}) display the same qualitative pattern with larger absolute OC, suggesting weaker internal context selection. Reasoning models diverge: \model{o3} is an outlier with near-ceiling $\Fonepivot$ on \dataset{TopiOCQA} and uniformly low OC, \model{claude-3.7-sonnet} and  \model{gemini-2.5-pro} follow with a similar trend and comparatively lower performance, whereas \model{deepseek-r1-32B} still struggles relatively in drawing crisp topic boundaries in persona-heavy settings, while showing a uniformly low OC. 

Overall, there are two key implications of the domain of the datasets. First, \emph{style matters}: in persona chat, the optimal policy is often to \emph{carry} unless explicitly signaled otherwise, whereas in knowledge-seeking QA the optimal policy is to \emph{reset} unless continuity is evident. Second, \emph{prompting matters}: lexical cue phrases (our \Vtwo{} setting) narrow the domain gap by externalizing the boundary decision. Cues help most where the training prior pushes in the wrong direction (e.g., \dataset{MSC}); they help less where the prior already aligns with the task (e.g., \dataset{TopiOCQA}). The same model can therefore look disciplined in one domain and challenged in another (e.g., \model{gpt-4o-mini} has low OC and solid $\Fonepivot$ on \dataset{TopiOCQA}, but high OC on \dataset{MSC}).

\subsection{Model specific failures}\label{app:modelfailures}

\paragraph{\model{o3}, \model{claude-3.7-sonnet} \& \model{gemini-2.5-pro} (reasoning, closed-source).}
On \dataset{TopiOCQA}, \model{o3} is near–ceiling on $\Fonepivot$ with the lowest overall $\OverCarry$, indicating precise boundary detection and disciplined shortlist selection and resetting at the pivot. On \dataset{MSC} it occasionally exhibits mild OC when the topic shift is subtle (e.g., persona or stance changes without topical lexical shift), consistent with a coherence prior that favors continuity in chit–chat. \model{o-3} is followed in performance by \model{claude-3.7-sonnet}, followed by \model{gemini-2.5-pro}, with almost consistent low $\OverCarry$ for all the models, and varying $\Fonepivot$ performance, indicating their detection capabilities along with reasoning. \textbf{Typical failure.} Missed \emph{fine–grained} pivots that are pragmatics–driven rather than topic–driven.

\paragraph{\model{deepseek-r1-32B} (reasoning, open–weights).}
Compared to \model{o3}, \model{deepseek-r1-32B} attains lower $\Fonepivot$ but remains relatively competitive on $\OverCarry$, suggesting that its step–by–step reasoning encourages relevance checks before carrying context. \textbf{Typical failure.} \emph{Boundary imprecision}: it often narrows to the right neighborhood but does not fire a PIVOT at $x_t=P$, yielding false REFINEs. 

\paragraph{\model{gpt-4o} (non–reasoning, closed–source).}
Strong $\Fonepivot$ and near–zero $\OverCarry$ on \dataset{TopiOCQA}; low OC on \dataset{MSC} shows very good context discipline relative to peers. The primary weakness is \emph{position bias}: later pivots in V3 show a measurable drop in PIVOT recall, especially when many coherent pre–pivot turns accumulate. 

\paragraph{\model{gpt-4o-mini} (non–reasoning, closed–source).}
Good $\Fonepivot$ overall, but high $\OverCarry$ on \dataset{MSC}: the model recognizes pivots yet continues to retrieve older persona turns. It shares the \emph{position bias} of \model{gpt-4o} but with larger magnitude. 

\paragraph{\model{gemma3-12B} \& \model{llama3.1-8B} (non-reasoning, open–weights).}
Moderate $\Fonepivot$ and clear improvements with cues, but consistently \emph{high} $\OverCarry$. A frequent pattern is \emph{label–context mismatch}: the model flags PIVOT yet still carries substantial pre–pivot turns in its shortlist. 

\paragraph{\model{phi4-14.7B} (non-reasoning, open–weights).}
Very low OC on \dataset{TopiOCQA} but under–detects pivots (many true PIVOTs labeled REFINE). The behavior suggests an \emph{over–aggressive reset policy} on context selection (leading to low OC) combined with a conservative labeler that hesitates to classify as PIVOT unless cues are explicit.

\paragraph{\model{mistral-7.2B} (non-reasoning, open–weights).}
Low $\Fonepivot$ and high $\OverCarry$ in our setting: by default the model exhibits a strong \emph{continuity prior}. It often treats concatenated sessions as one thread unless given overt boundary signals.

\subsection{Improvements to models}\label{app:improvements}

We offer training–time learning signals and inference–time system design recommendations that directly target the observed failure modes (position bias, label–shortlist mismatch, cue over–reliance, and domain–based continuity priors).

\paragraph{Training–time data and objectives for open models.}
\begin{enumerate}[leftmargin=*]

\item \textbf{Instruction Tuning for boundary awareness.} Extend SFT with samples that (i) lack lexical cues but require PIVOT and (ii) require empty carry on PIVOT. Short, explicit rubrics (``On PIVOT, \emph{do not carry} prior turns'') help close the label–shortlist gap.

\item \textbf{Late–pivot curriculum.} To reduce position bias, oversample sessions with large refine–prefix upto $P$ (heavy–tail over $P$) and include near–pivot \emph{hard negatives}. This schedule would help in improving robustness to late pivots.

\item \textbf{Cue diversity and ablations.} Mix cue/no–cue data (and paraphrastic cues) so the model learns pivots as a semantic property rather than a token separator. At the same time, keep a small held–out cue set to check over–fitting to specific phrases.
\end{enumerate}

\paragraph{Inference Time System design}

\begin{enumerate} 

\item \textbf{Prompt Engineering with Boundary awareness.} Chain-of-thought style instructions in the prompt along with a large number of few-shot in-context examples for challenging boundary cases can improve the inference time performance of the out-of-the-box models for both detection and context resetting tasks. Similarly, including challenging examples of late-pivot detections can help reduce the degradation in performance, as context grows.

\item \textbf{Two Stage Controller.} We can modify the system to decompose the prediction problem into multiple steps. 

\begin{enumerate}[leftmargin=*]
\item \textbf{Stage–1 pivot gate (binary).} Train a compact classifier $p_\theta(y_t\!\in\!\{\text{PIVOT},\text{REFINE}\}\mid h_t)$ on top of hidden summaries $h_t$ (e.g., final token CLS for the current turn plus a pooled history embedding) to trade off pivot recall vs.\ false positives.

\item \textbf{Stage–2 context selector (conditional).} Given $\hat{y}_t$, select shortlist $S_t$: if $\hat{y}_t=\text{PIVOT}$, \emph{enforce} $S_t=\varnothing$ (hard rule and can be done programmatically); if $\hat{y}_t=\text{REFINE}$, select from the current thread only (e.g., $k$ most recent same–segment turns by a relevance scorer). 
\end{enumerate}
\end{enumerate}

\paragraph{Reasoning models: structure the process.}
For chain–of–thought or tool–use models, separate \emph{Plan} and \emph{Answer}: (i) decide \textsc{Pivot/Refine} with a brief justification (``boundary rationale''), (ii) \emph{then} generate the answer using only the approved shortlist.

\paragraph{Mitigating position bias (\texorpdfstring{$P$}{P}–aware control).}
Later pivots (as $P\!\uparrow$) are harder to detect due to anchoring on context and latent retrieval. To solve this problem, we can treat the turns within a session differently, as the number of turns within a session grows. Specifically, we can add: (i) \emph{recency–biased context windows} (summarize early turns into a single synopsis), (ii) a \emph{$P$–aware prior} that linearly increases the pivot probability with depth, (iii) \emph{explicit anti–anchoring prompts} (``If this is a new topic, ignore prior messages entirely''), and (iv) \emph{cue amplification} for large $P$ (stronger lexical boundary markers when long histories accumulate).

\section{Limitations}\label{app:limitations}
There are several limitations of the current study which we address and acknowledge in this section. 

\paragraph{Scope of our work.}
Our study isolates two behaviors: (i) recognizing topic shifts (\emph{PIVOT}) versus within-topic continuation (\emph{REFINE}), and (ii) deciding which prior turns to carry forward as a \emph{shortlist}. We do \emph{not} evaluate downstream task quality (e.g., answer correctness in QA) or long-form generation faithfulness, to focus on context understanding as the primary problem. As a result, a model may correctly flag a \emph{PIVOT} yet still produce an incorrect answer; conversely, a model may answer well on \emph{REFINE} turns while mislabeling the turn type. The interplay between task success (answer quality) as well as interaction competence (identifying $\pivot$ or $\refine$ and \emph{context} resetting is an interesting research problem worth investigating further. Additionally, we restrict our study to identical \emph{zero-shot} prompts across models to evaluate the \emph{inherent} capabilities of out-of-the-box LLMs. Advanced prompting strategies such as Chain-of-thought prompting and including in-context examples in the prompt could require model specific modifications and are left as future work. 

\paragraph{Over-carry is measured at the pivot; under-carry on refines is more challenging to measure.}
Our primary behavioral error is \emph{over-carry}, i.e. models retaining irrelevant context at a true \emph{PIVOT}. (See Section \ref{app:metrics} for details). While we also report the detection $\Fonepivot$ on \emph{REFINE} turns as a measure of detection performance, observing true \emph{under-carry} performance is challenging since the labeling for a ground-truth shortlist of relevant context for a \emph{REFINE} turn may be unavailable. This may be inferred through different techniques which would require addressing challenges such as ambiguity, coreference, dialogue flow, capturing nuances, etc and approximate the human relevance of context required for responding to a current turn. This would also require human validation or annotation for a subset of the turns to calibrate the shortlist quality and quantify any bias.

\paragraph{Binary turn typing omits \emph{partial} or \emph{soft} pivots.}
We limit turn types to a binary \{\emph{REFINE} or \emph{PIVOT}\} label, however, natural conversations exhibit graded topical drift and blended intents, long studied in topic segmentation \citep{hearst1997text,eisenstein2008bayesian,galley2003discourse}. Our construction treats any head splice from a new session as a hard \emph{PIVOT}. In practice, users often pivot partially where a second request introduces a new intent while remaining anchored to the prior entity. Our binary labels would count this as \emph{PIVOT}, although a graded scheme (or multi-label \emph{intent}+\emph{topic} labeling) may be more faithful for such \emph{soft} pivots.

\paragraph{Domain and language coverage.}
We evaluate on \dataset{TopiOCQA} (Wikipedia-oriented QA) and \dataset{MSC} (long-term persona chat). Both are in English; we do not test multilingual pivot detection, code-switched languages, or  chats in specialized domains (e.g., legal or medical). Moreover, topic familiarity (e.g., popular entities in \dataset{TopiOCQA}) can interact with pivot detection by making carry decisions easier \citep{rajpurkar2016squad,anantha2020open}. Extending to more domains and languages is left to future work.

\section{Statements}\label{app:statements}

\paragraph{Ethics Statement.}
Our study does not involve any human subjects. We also do not foresee any negative societal impacts, discrimination/bias/fairness concerns, and privacy or security concerns from the outcomes of our research. 

\paragraph{Reproducibility Statement.}
All of our experiments were conducted with fixed seeds for ensuring reproducibility.

\paragraph{Statement on LLM Usage.}
While LLM-based coding assistants were used in the process of conducting experiments, we did not use any LLMs for writing of the paper (including the main text, Appendix and References).

\end{document}